%% file: main.tex
\documentclass[10pt,twocolumn,letterpaper]{article}

\usepackage[pagenumbers]{iccv}              

\usepackage{amsmath}
\usepackage{amssymb}
\usepackage{enumitem}
\usepackage{natbib}
\usepackage{graphicx}
\usepackage{float}
\usepackage{physics}
\usepackage{nth}
\usepackage{soul}
\usepackage{xcolor}


\definecolor{iccvblue}{rgb}{0.21,0.49,0.74}
\usepackage[pagebackref,breaklinks,colorlinks,allcolors=iccvblue]{hyperref}

\def\paperID{20} 
\def\confName{ICCV}
\def\confYear{2025}

\title{AASeg: Attention Aware Network for Real Time Semantic Segmentation}

\author{Abhinav Sagar\\
University of Maryland, College Park, Maryland\\
College Park, Maryland\\
{\tt\small asagar@umd.edu}
}

\begin{document}

\maketitle

\begin{abstract}

Semantic segmentation is a fundamental task in computer vision that involves dense pixel-wise classification for scene understanding. Despite significant progress, achieving high accuracy while maintaining real-time performance remains a challenging trade-off, particularly for deployment in resource-constrained or latency-sensitive applications. In this paper, we propose AASeg, a novel Attention-Aware Network for real-time semantic segmentation. AASeg effectively captures both spatial and channel-wise dependencies through lightweight Spatial Attention (SA) and Channel Attention (CA) modules, enabling enhanced feature discrimination without incurring significant computational overhead. To enrich contextual representation, we introduce a Multi-Scale Context (MSC) module that aggregates dense local features across multiple receptive fields. The outputs from attention and context modules are adaptively fused to produce high-resolution segmentation maps. Extensive experiments on Cityscapes, ADE20K, and CamVid demonstrate that AASeg achieves a compelling trade-off between accuracy and efficiency, outperforming prior real-time methods. 

\end{abstract}

\section{Introduction}

Semantic segmentation is a core problem in computer vision, where the goal is to assign a semantic label to every pixel in an image. This task plays a vital role in various real-world applications such as autonomous driving, robotic navigation, augmented reality, and medical imaging. However, despite recent advancements in deep learning-based segmentation models, achieving an optimal balance between segmentation accuracy and inference speed remains a significant challenge, particularly in resource-constrained or latency-critical environments.

Early work, such as the Fully Convolutional Network (FCN) \citep{long2015fully} demonstrated that convolutional neural networks could be extended to pixel-level prediction tasks. While FCNs achieved strong semantic representation, they struggled with accurately capturing fine-grained details, especially around object boundaries. To address this, later works employed atrous (dilated) convolutions \citep{yu2015multi, chen2017rethinking} to expand the receptive field without downsampling, thereby improving context aggregation. However, this enhancement often came with a substantial increase in computational overhead, making such methods less suitable for real-time deployment.

In response to the growing demand for efficient segmentation models, several lightweight architectures have been proposed. SegNet \citep{badrinarayanan2017segnet} introduced an encoder–decoder structure with skip connections for improved inference speed. Networks like ENet \citep{paszke2016enet}, ESPNet \citep{mehta2018espnet}, and Fast-SCNN \citep{poudel2019fast} further pushed the boundary of real-time semantic segmentation through architectural simplification, depth-wise separable convolutions, and efficient encoder-decoder designs. While these models improve throughput (measured in frames per second, FPS), they often do so at the cost of degraded segmentation accuracy, particularly in challenging scenarios involving small objects or complex spatial layouts.

To address these limitations, we propose AASeg (Attention-Aware Segmentation Network), a real-time semantic segmentation framework that explicitly enhances both spatial and channel-wise feature representations using attention mechanisms. Our network introduces three key components: (1) a Spatial Attention (SA) module to emphasize salient regions in the spatial domain, (2) a Channel Attention (CA) module to reweight informative channels adaptively, and (3) a Multi-Scale Context (MSC) module to aggregate dense local context across varying receptive fields without significant computational burden. By integrating these modules into a lightweight yet expressive architecture, AASeg achieves high segmentation accuracy while maintaining fast inference speeds suitable for real-time applications.

\begin{figure}[htp]
\centering
\includegraphics[width=8cm]{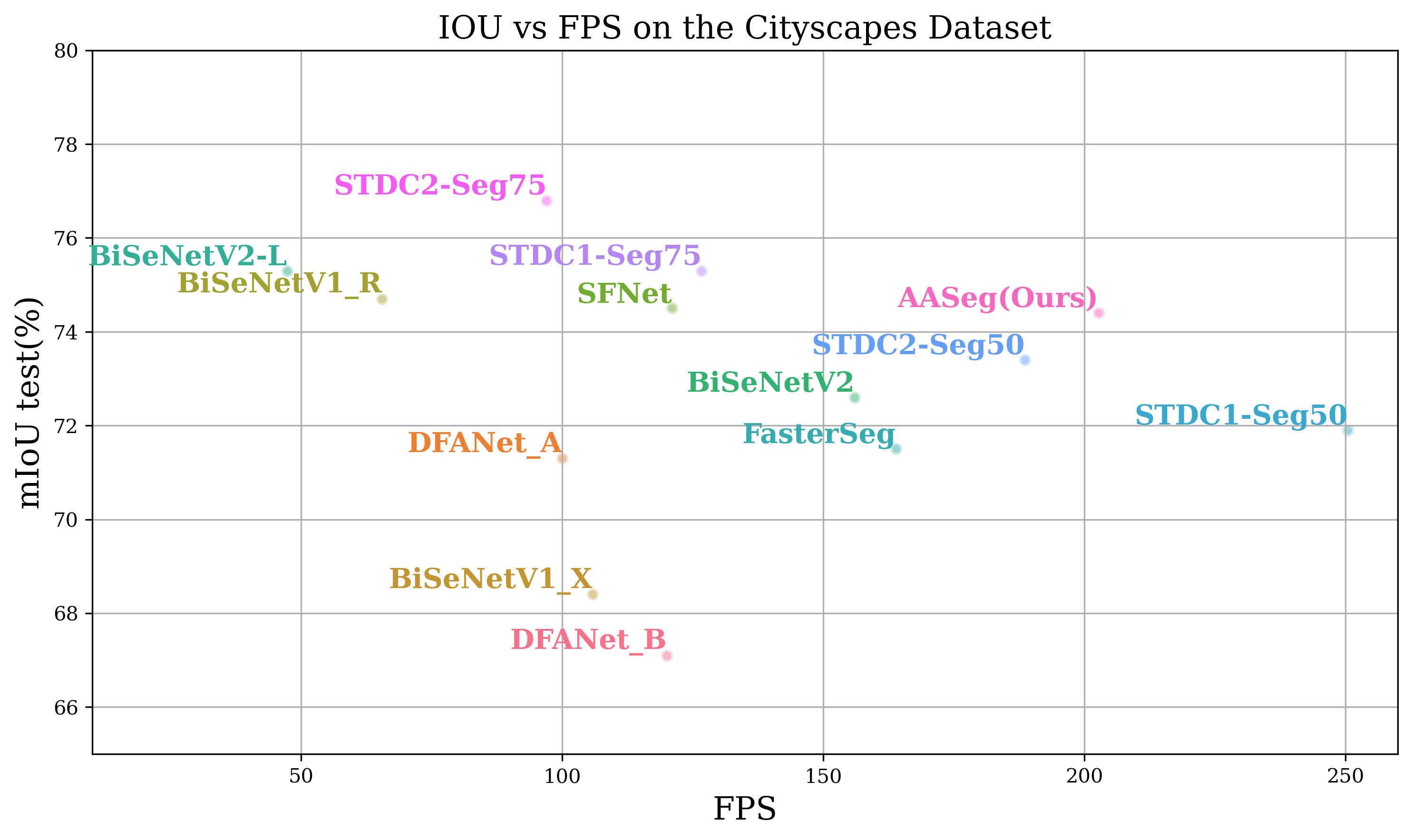}
\caption{Comparison of speed vs. accuracy on the Cityscapes test set. AASeg achieves a compelling balance, delivering high segmentation accuracy while maintaining real-time performance.}
\label{speed_accuracy}
\end{figure}

We evaluate AASeg on standard benchmarks including Cityscapes, ADE20K, and CamVid, and show that it outperforms most existing real-time methods. Specifically, AASeg achieves 74.4\% mIoU on the Cityscapes test set while operating at 202.7 FPS, setting a new benchmark for high-speed semantic segmentation. The speed–accuracy trade-off of state-of-the-art semantic segmentation methods is illustrated in \autoref{speed_accuracy}, which presents results on the Cityscapes test set. Our proposed method, AASeg, achieves a higher mean Intersection-over-Union (mIoU) while maintaining superior inference speed (FPS), outperforming most existing real-time approaches.

\section{Related Work}

Real-time semantic segmentation has gained significant attention due to its crucial role in safety-critical applications such as autonomous driving and robotic perception. Recent advances aim to optimize the trade-off between segmentation accuracy and computational efficiency. We categorize related work into four key areas: attention mechanisms, lightweight backbones, multi-resolution and multi-path designs, and contextual feature aggregation.

\subsection{Attention Mechanisms}

Attention modules have proven effective in improving feature representation by modeling long-range dependencies. The Dual Attention Network (DANet) \citep{fu2019dual} introduces spatial and channel attention modules to adaptively enhance the receptive field, leading to more discriminative features. Similarly, Squeeze-and-Excitation Networks \citep{iandola2016squeezenet} apply channel-wise recalibration to focus on the most informative feature channels, improving both accuracy and efficiency. These attention-based designs motivate our use of spatial and channel attention to enhance feature encoding in a lightweight framework.

\subsection{Lightweight Backbones and Efficient Convolutions}

To reduce computation, several approaches design compact network architectures. ENet \citep{paszke2016enet} aggressively prunes convolutional filters and applies early downsampling to minimize latency. ESPNet \citep{mehta2018espnet} introduces an Efficient Spatial Pyramid (ESP) module to decompose standard convolutions into parallel dilated convolutions, enabling fast and accurate segmentation. DFA-Net \citep{li2019dfanet} employs a lightweight backbone and depth-wise separable convolutions along with multi-scale feature fusion to balance performance and speed.

\subsection{Multi-Resolution and Multi-Path Architectures}

Multi-branch architectures are effective in capturing both global semantics and fine spatial details. ICNet \citep{zhao2018icnet} proposes a cascade framework with high, medium, and low-resolution branches to progressively refine segmentation. BiSeNet \citep{yu2018bisenet} introduces a two-path structure consisting of a Spatial Path (SP) for preserving spatial resolution and a Context Path (CP) for enlarging the receptive field. BiSeNetV2 \citep{yu2020bisenet} further refines this concept using a multi-path fusion strategy to improve both accuracy and efficiency.

\subsection{Contextual Feature Aggregation}

Several works leverage contextual information to improve prediction quality. SwiftNet \citep{orsic2019defense} uses a ResNet-based encoder with lateral connections to recover spatial details at high speed. RefineNet \citep{lin2017refinenet} employs multi-path refinement to progressively enhance feature maps, but relies heavily on deep backbones and neglects global context efficiency.

Our main contributions are summarized as follows:

\begin{itemize}
    
\item We propose AASeg, a novel Attention-Aware Network designed specifically for real-time semantic segmentation, targeting the optimal balance between accuracy and efficiency.

\item AASeg introduces a modular architecture consisting of three core components: a Spatial Attention (SA) module that models spatial dependencies to emphasize informative regions, a Channel Attention (CA) module that adaptively reweights feature channels to capture discriminative semantic features, and a Multi-Scale Context (MSC) module that aggregates contextual information across multiple receptive fields to enhance hierarchical feature representation.

\item We conduct extensive experiments and ablation studies on three benchmark datasets—Cityscapes, CamVid, and ADE20K—demonstrating that AASeg not only improves segmentation accuracy but also significantly boosts inference speed compared to most other real-time methods.

\end{itemize}

\section{Background}

\subsection{Spatial Information}

Preserving spatial information is critical in semantic segmentation, as it directly impacts the ability to accurately delineate object boundaries and fine-grained structures. Modern segmentation frameworks incorporate spatial encoding mechanisms to retain this information throughout the network. For instance, Dilated Convolutions, as used in DUC \citep{wang2018understanding}, PSPNet \citep{zhao2017pyramid}, and DeepLab v2 \citep{chen2017rethinking}, allow for enlarging the receptive field without reducing the spatial resolution of feature maps. This approach helps preserve detailed spatial cues crucial for high-precision predictions.

\subsection{Context Information}

In addition to spatial details, capturing global and local context is essential for understanding object relationships and scene structure. Contextual information enables the network to differentiate between visually similar regions based on surrounding cues. To this end, several methods adopt multi-scale feature extraction strategies. For example, ASPP (Atrous Spatial Pyramid Pooling) in DeepLab \citep{chen2017rethinking} captures features at multiple scales using dilated convolutions with varying dilation rates. Similarly, PSPNet \citep{zhao2017pyramid} introduces a Pyramid Pooling Module (PPM) that performs average pooling at different scales to aggregate global context. These methods enhance the model's ability to recognize objects under varying spatial extents and scene complexities.

\subsection{Attention Mechanisms}

Attention mechanisms have been increasingly adopted in segmentation networks to selectively enhance relevant features while suppressing redundant ones. Squeeze-and-Excitation Networks \citep{hu2018squeeze} pioneered the use of channel-wise attention, significantly improving performance in image recognition tasks. For semantic segmentation, non-local attention and contextual attention approaches have proven effective. For example, \citep{yu2018learning} proposed an attention module that refines feature representations by modeling global dependencies.

\subsection{Feature Fusion}

Feature fusion plays a pivotal role in combining different sources of information, particularly the integration of low-level spatial details and high-level semantic context. Architectures such as BiSeNet utilize a dual-path design to separately capture spatial and contextual features, which are then fused to produce the final output. Effective fusion strategies have also been applied to related tasks, including classification, detection, and instance segmentation. In the context of semantic segmentation, proper fusion allows the network to benefit from both sharp localization and robust semantic understanding.

\section{Proposed Method}

\subsection{Dataset}

The following datasets have been used to benchmark our results:

\begin{itemize}

\item Cityscapes: It is used for urban street segmentation. The 5000 annotated images are used in our experiments, which are divided into 2975, 500, and 1525
images for training, validation, and testing, respectively.

\item ADE20K: This dataset contains labels of 150 object categories. The dataset includes 20k,2k, and 3k images for training, validation, and testing, respectively.

\item CamVid: This dataset is used for semantic segmentation for autonomous driving scenarios. It is composed of 701 densely annotated images.

\end{itemize}

\subsection{Network Architecture}

The primary goal of semantic segmentation is to map an input RGB image $X \in \mathbb{R}^{H \times W \times 3}$ to a semantic label map $Y \in \mathbb{R}^{H \times W \times C}$, where $H$ and $W$ are the image dimensions and $C$ is the number of semantic classes. The input image $X$ is transformed into hierarchical feature representations ${F_l}_{l=1}^3$, where $F_l \in \mathbb{R}^{H_l \times W_l \times C_l}$ is the feature map from the $l^\text{th}$ stage.

Unlike many existing segmentation frameworks that rely on pre-trained backbones (e.g., ResNet, MobileNet), our proposed architecture, AASeg, is designed from scratch with lightweight modules optimized for real-time performance. The input image is initially passed through a convolutional block consisting of a convolutional layer, batch normalization, and ReLU activation.

We define a convolutional operation $W^n(x)$ as:

\begin{equation}
W^n(x) = \mathbf{W}^{n \times n} \odot x + \mathbf{b}
\end{equation}

where $\odot$ denotes convolution, $\mathbf{W}^{n \times n}$ is an $n \times n$ kernel, $x$ is the input feature map, and $\mathbf{b}$ is the bias term.

\subsubsection{Multi-Scale Context (MSC) Module}

To capture rich local and global context, we introduce a Multi-Scale Context (MSC) module. The input feature map is processed using convolutions of varying kernel sizes: $1 \times 1$, $3 \times 3$, and $5 \times 5$. These outputs are fused and then passed through a $1 \times 1$ convolution to reduce dimensionality from 2048 to 256, producing an output of size $H \times W \times n_c$.

Subsequently, a series of dilated convolutions with increasing dilation rates ($d = 3, 6, 12$) are applied. The input at each stage is a concatenation of the original feature map and all preceding outputs. Finally, the three dilated outputs are concatenated with the original feature map to enhance the context representation.

\subsubsection{Spatial Attention (SA) Module}

The Spatial Attention (SA) module captures spatial dependencies by focusing on relevant regions of the feature map. It is defined as:

\begin{equation}
f_{\text{SA}}(x) = \sigma(W_2(\text{ReLU}(W_1(x))))
\end{equation}

where $W_1$ and $W_2$ are $1 \times 1$ convolution layers, $\sigma$ denotes the sigmoid function, and ReLU is the activation function. This module generates a spatial attention mask that enhances informative locations in the feature map and is shown in \autoref{fig:sa_module}:

\begin{figure}[htp]
\centering
\includegraphics[width=6cm]{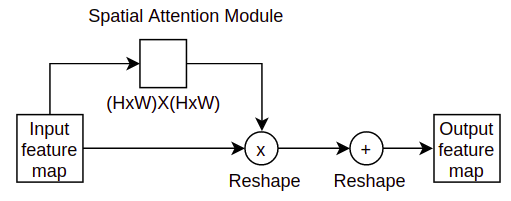}
\caption{Details of our Spatial Attention (SA) module.}
\label{fig:sa_module}
\end{figure}

\subsubsection{Channel Attention (CA) Module}

To capture inter-channel dependencies and semantic saliency, we employ a Channel Attention (CA) module. It is defined as:

\begin{equation}
f_{\text{CA}}(x) = \sigma(W_2(\text{ReLU}(W_1(\text{AvgPool}(x)))))
\end{equation}

Here, $\text{AvgPool}$ denotes global average pooling, and $W_1$, $W_2$ are $1 \times 1$ convolution layers. The resulting attention map is used to modulate the input feature map channel-wise and is shown in \autoref{fig:ca_module}:

\begin{figure}[htp]
\centering
\includegraphics[width=6cm]{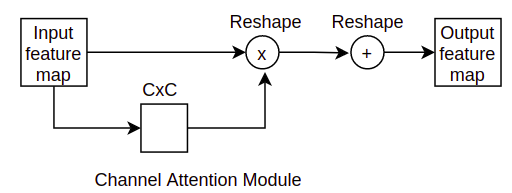}
\caption{Details of our Channel Attention (CA) module.}
\label{fig:ca_module}
\end{figure}

\subsubsection{Feature Aggregation}

To integrate the outputs of the SA, CA, and MSC modules, we define a unified aggregation operation. Let $\oplus$ denote the concatenation operator. The intermediate concatenated feature $x_{\text{concat}}$ is defined as:

\begin{equation}
x_{\text{concat}} = x_1 \oplus x_2 \oplus x_3
\end{equation}

The output of the AASeg module is then computed as:

\begin{equation}
\begin{split}
x_{\text{AASeg}} = & \left( f_{\text{SA}}(x_{\text{concat}}) \otimes x_{\text{concat}} \right) \oplus \\
                   & \left( f_{\text{CA}}(x_{\text{concat}}) \otimes x_{\text{concat}} \right) \oplus \\
                   & \left( f_{\text{MSC}}(x_{\text{concat}}) \otimes x_{\text{concat}} \right)
\end{split}
\end{equation}

Here, $\otimes$ represents element-wise multiplication.

\subsubsection{Convolutional Blocks and Output}

We denote the output of the $i^\text{th}$ convolutional block as:

\begin{equation}
x_i = \text{ConvBlock}i(x{i-1}, k_i)
\end{equation}

where each $\text{ConvBlock}_i$ consists of a convolutional layer with kernel size $k_i$, batch normalization, and ReLU activation.

To produce the final output, we apply a fusion operation that integrates all hierarchical outputs:

\begin{equation}
x_{\text{output}} = F(x_1, x_2, ..., x_n)
\end{equation}

\subsubsection{Overall Architecture}

\begin{figure*}[htp]
\centering
\includegraphics[width=16cm]{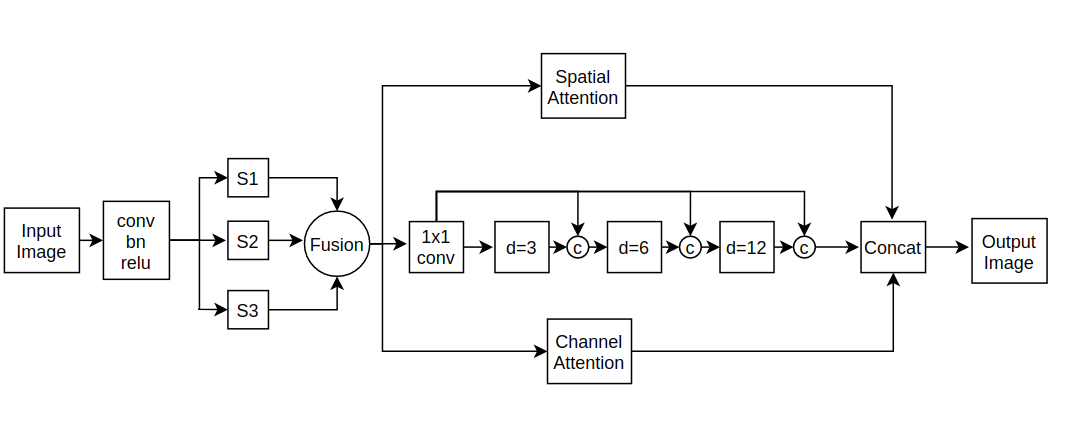}
\caption{Overview of the proposed AASeg architecture. “c” denotes concatenation.}
\label{fig:architecture}
\end{figure*}

The full AASeg architecture is depicted in Figure~\ref{fig:architecture}, illustrating the flow from input through spatial, channel, and context attention modules to the final segmentation prediction.

\subsection{Loss Functions}

We use the cross-entropy loss function to weigh the difference between the forward propagation result of the network and the ground truth of the samples. The cross-entropy loss is calculated as:

\begin{equation}
{L}_{c e}=\frac{1}{N} \sum_{n=1}^{N}\left[y_{n} \log \hat{y}_{n}+\left(1-y_{n}\right) \log \left(1-\hat{y}_{n}\right)\right]
\end{equation}

Where $N$ denotes the total number of samples, $y_{n}$ denotes the probability that the forward propagation result is true, and 1-$y_{n}$ denotes the probability that the forward propagation result is false.

We also use the auxiliary supervision loss $L_{aux}$ to improve the model's performance, thus making it easier to
optimize. The auxiliary loss can be defined as:

\begin{equation}
\begin{split}
\mathcal{L}_{aux} = -\frac{1}{B N} \sum_{i=1}^{B} \sum_{j=1}^{N} \sum_{k=1}^{K} 
\mathbb{I}\left(g_{j}^{i} = k\right) \\
\times \log \left(
\frac{\exp \left(p_{j, k}^{i}\right)}{\sum_{m=1}^{K} \exp \left(p_{j, m}^{i}\right)}
\right)
\end{split}
\end{equation}

\begin{equation}
{I}\left(g_{j}^{i}=k\right)=\left\{\begin{array}{ll}
1, & g_{j}^{i}=k \\
0, & { otherwise }
\end{array}\right.
\label{igi}
\end{equation}

where $B$ is the mini batch size, $N$ is the number of pixels
in every batch; $K$ is the number of categories; $p_{ijk}$ is the prediction of the $j^{th}$ pixel in the $i^{th}$
sample for the $k^{th}$ class, $I(g_{ij} = k)$ is a function which is defined in \autoref{igi}.

The class attention loss $L_{cls}$ from the channel attention module is also used. The class Attention loss is defined as follows:

\begin{equation}
\begin{split}
\mathcal{L}_{cls} = -\frac{1}{B N} \sum_{i=1}^{B} \sum_{j=1}^{N} \sum_{k=1}^{K} 
\mathbb{I}\left(g_{j}^{i} = k\right) \\
\times \log \left(
\frac{\exp \left(a_{j, k}^{i}\right)}{\sum_{m=1}^{K} \exp \left(a_{j, m}^{i}\right)}
\right)
\end{split}
\end{equation}

where $a_{ijk}$ is the value generated by the class attention map
of the $j^{th}$ pixel in the $i^{th}$ sample for the $k^{th}$ class. We combine the three terms to balance the final loss term as
follows:

\begin{equation}
\mathcal{L}=\lambda_{1} \mathcal{L}_{c e}+\lambda_{2} \mathcal{L}_{c l s}+\lambda_{3} \mathcal{L}_{a u x}
\end{equation}

Where $\lambda1$, $\lambda2$, and $\lambda3$ are set as 1, 0.5, and 0.5 to balance the loss. 

\subsection{Implementation Details}

The PyTorch deep learning framework is used to carry out our experiments. We use stochastic gradient descent (SGD) as optimizer, batch size of 16, momentum of 0.9, and weight decay of $5e^{-4}$. The network is trained for 20K iterations with an initial learning rate value of 0.01. The “poly” learning rate policy is used to decay the initial learning rate while training the model to reduce over-fitting. Data augmentation operations like random horizontal flip, random resizing with a scale range of [1.0, 2.0], and random cropping with a crop size of $1024 \times 1024$ were done.  

\subsection{Evaluation Metrics}

For quantitative evaluation of our network's performance, we employ multiple standard metrics to comprehensively assess both accuracy and efficiency. The primary accuracy metric is the mean Intersection-over-Union (mIoU), which calculates the average class-wise overlap between the predicted segmentation and the ground truth, providing a robust measure of segmentation quality. To evaluate computational efficiency, we report the number of floating-point operations (FLOPs), reflecting the computational complexity of the model, and frames per second (FPS), indicating the real-time processing speed of the network during inference. Together, these metrics offer a balanced benchmarking framework that captures both the effectiveness and practicality of our method.

\section{Results and Discussion}

We present the segmentation accuracy and inference speed of our proposed Attention Aware Network (AASeg) on the Cityscapes validation and test sets in \autoref{table1}. The model is trained using both the Cityscapes training and validation sets before submitting predictions to the official Cityscapes online evaluation server. \autoref{table1} compares AASeg with several recent state-of-the-art semantic segmentation methods in terms of mean Intersection-over-Union (mIoU) on both validation and test sets, input resolution, backbone architecture, and inference speed measured in frames per second (FPS). Our AASeg model achieves a competitive mIoU of 74.8\% on the validation set and 74.4\% on the test set without relying on any backbone network, demonstrating that our design effectively extracts robust features despite the absence of a large pretrained encoder. In terms of speed, AASeg processes images at 202.7 FPS at a resolution of $512 \times 1024$, outperforming many existing models with backbones, including BiSeNetV2 and STDC variants, thereby offering a superior speed-accuracy tradeoff for real-time applications.
Compared to lightweight models like BiSeNetV2 and FasterSeg, AASeg provides higher segmentation accuracy while maintaining competitive inference speed. Although some models, such as STDC1-Seg50, achieve higher FPS, they exhibit lower mIoU, indicating that AASeg offers a more balanced performance for practical deployment. The results highlight the effectiveness of our spatial-channel attention modules combined with the multi-scale context module in enhancing feature representation and segmentation precision without incurring additional computational overhead.

\begin{table*}[h]
  \caption{Comparisons with other state-of-the-art methods using Cityscapes dataset. No indicates the method does not have any backbone for training and testing. Best results are highlighted in bold.}
  \label{table1}
  \centering
  \begin{tabular}{llllll}
  \toprule
    Model &Resolution &Backbone &mIoU val(\%) &mIoU test(\%) &FPS\\
   \midrule
DFANet B &1024 $\times$ 1024 &Xception B &- &67.1 &120\\
DFANet A &1024 $\times$ 1024 &Xception A &- &71.3 &100\\
BiSeNetV1 &768 $\times$ 1536 &Xception39 &69.0 &68.4 &105.8\\
BiSeNetV1 &768 $\times$ 1536 &ResNet18 &74.8 &74.7 &65.5\\
SFNet &1024 $\times$ 2048 &DF1 &- &74.5 &121\\
BiSeNetV2 &512 $\times$ 1024 &no &73.4 &72.6 &156\\
BiSeNetV2-L &512 $\times$ 1024 &no &75.8 &75.3 &47.3\\
FasterSeg &1024 $\times$ 2048 &no &73.1 &71.5 &163.9\\
STDC1-Seg50 &512 $\times$ 1024 &STDC1 &72.2 &71.9 &\textbf{250.4}\\
STDC2-Seg50 &512 $\times$ 1024 &STDC2 &74.2 &73.4 &188.6\\
STDC1-Seg75 &768 $\times$ 1536 &STDC1 &74.5 &75.3 &126.7\\
STDC2-Seg75 &768 $\times$ 1536 &STDC2 &\textbf{77.0} &\textbf{76.8} &97.0\\
AASeg &512 $\times$ 1024 &no &74.8 &74.4 &202.7\\
    \bottomrule
  \end{tabular}
\end{table*}

The qualitative results on the Cityscapes validation set are presented in \autoref{results}:

\begin{figure*}
    \centering
    \includegraphics[width=12cm]{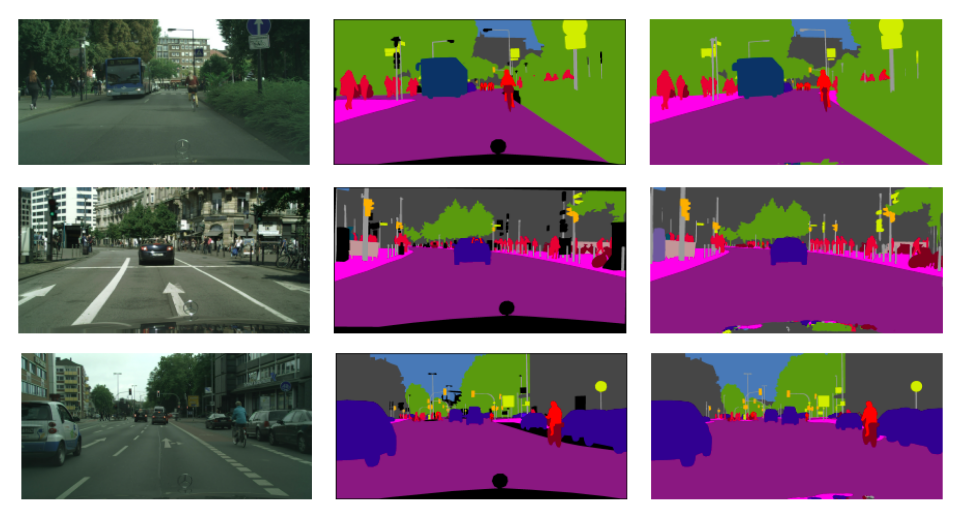}
    \caption{Visualized segmentation results on Cityscapes validation set. The three columns left-to-right refer to the input image, ground truth, and prediction from our network.}
    \label{results}
\end{figure*}

We evaluate the performance of our proposed AASeg model on the CamVid dataset. The comparative results with recent state-of-the-art methods are summarized in \autoref{table2}. Our experiments use an input resolution of $720 \times 960$ pixels, consistent with previous works to ensure a fair comparison. As shown in \autoref{table2}, AASeg achieves a high mean Intersection-over-Union (mIoU) score of 73.5\%, which is competitive with other leading methods such as BiSeNetV2-L and SFNet variants that employ pretrained backbones. Notably, our model does not rely on any backbone network, yet it manages to deliver comparable or superior accuracy. In terms of inference speed, AASeg processes frames at 188.7 FPS, demonstrating its suitability for real-time applications in urban scene understanding and autonomous driving scenarios. This speed outperforms most backbone-based models, including SFNet and BiSeNetV1 with ResNet18, while closely approaching the fastest model, STDC1-Seg, which reaches 197.6 FPS but with slightly lower mIoU. 

\begin{table*}
  \caption{Comparisons with other state-of-the-art methods on the CamVid dataset. No indicates the method does not have a backbone for both training and testing. Best results are highlighted in bold.}
  \label{table2}
  \centering
  \begin{tabular}{lllll}
  \toprule
    Model &Resolution &Backbone &mIoU(\%) &FPS\\
   \midrule
DFANet A &720 $\times$ 960 &no &64.7 &120\\
DFANet B & 720 $\times$ 960 &no &59.3 &160\\   
BiSeNetV1 &720 $\times$ 960 &Xception39 &65.6 &175\\
BiSeNetV1 &720 $\times$ 960 &ResNet18 &68.7 &116.3\\
BiSeNetV2 &720 $\times$ 960 &no &72.4 &124.5\\
BiSeNetV2-L &720 $\times$ 960 &no &73.2 &32.7\\
SFNet &720 $\times$ 960 &DF2 &70.4 &134.1\\
SFNet &720 $\times$ 960 &ResNet-18 &73.8 &35.5\\
SFNet &720 $\times$ 960 &DF2 &70.4 &134.1\\
SFNet & 720 $\times$ 960 &ResNet-18 &73.8 &35.5\\
STDC1-Seg &720 $\times$ 960 &STDC1 &73.0 &\textbf{197.6}\\
STDC2-Seg &720 $\times$ 960 &STDC2 &\textbf{73.9} &152.2\\
AASeg &720 $\times$ 960 &no &73.5 &188.7\\
    \bottomrule
  \end{tabular}
\end{table*}

We further validate the effectiveness of our proposed AASeg model, \autoref{table3} presents a performance comparison between AASeg and several recent state-of-the-art semantic segmentation methods evaluated on the ADE20K validation set. Our model achieves a mean Intersection-over-Union (mIoU) of 46.29\%, outperforming well-established approaches such as PSPNet and SFNet by a noticeable margin. Specifically, AASeg surpasses PSPNet101 by nearly 3\% and also improves upon SFNet variants, demonstrating the strong representational power of our attention-aware architecture. Also, AASeg requires only 80.26 GFLOPs, which is significantly lower than PSPNet50 and PSPNet101 and only slightly above the most efficient SFNet variant with 75.7 GFLOPs. This balance of high accuracy and moderate computational complexity highlights AASeg’s potential for real-world applications, especially in resource-constrained environments.

\begin{table}[h]
  \caption{Performance comparison of results reported on the ADE20K validation set. Best results are highlighted in bold.}
  \label{table3}
  \centering
  \begin{tabular}{lll}
  \toprule
Method &Mean IoU(\%) &GFLOPs\\
   \midrule
PSPNet50 &42.78 &167.6\\
PSPNet101 &43.29 &238.4\\
SFNet &42.81 &\textbf{75.7}\\
SFNet &44.67 &94.0\\
DCANet &45.49 &-\\
AASeg &\textbf{46.29} &80.26\\
    \bottomrule
  \end{tabular}
\end{table}

\section{Ablation Study}

\subsection{Ablation Study on Upsampling Methods}

We investigate the impact of different upsampling operations on the performance of our AASeg network. Specifically, we compare three commonly used techniques: bilinear upsampling, deconvolution (transposed convolution), and nearest neighbor upsampling. The results of this ablation study on the Cityscapes validation set are summarized in \autoref{ablation1}. Our experiments show that bilinear upsampling achieves the best performance with a mIoU of 79.2\%, slightly outperforming both deconvolution (78.5\%) and nearest neighbor upsampling (78.3\%). Although the differences are relatively small, bilinear upsampling provides a good trade-off between accuracy and computational simplicity. Deconvolution, while being a learnable upsampling method, does not significantly improve the segmentation accuracy in our setup, possibly due to added complexity or training instability. Nearest neighbor upsampling, the simplest and fastest method, also performs competitively but slightly trails the other two approaches. Based on these findings, bilinear upsampling is chosen as the default upsampling method in our network, as it balances accuracy, smoothness in feature maps, and computational efficiency. This insight reinforces the notion that simple interpolation-based upsampling methods can be effective for dense prediction tasks when combined with strong attention mechanisms.

\begin{table}[h]
  \caption{Ablation study with Upsampling
operation in our network using the Cityscapes validation set. Best results are highlighted in bold.}
  \label{ablation1}
  \centering
  \begin{tabular}{ll}
  \toprule
Method &mIoU (\%)\\
   \midrule
bilinear upsampling &\textbf{79.2}\\
deconvolution &78.5\\
nearest neighbor &78.3\\
    \bottomrule
  \end{tabular}
\end{table}

\subsection{Ablation Study on Kernel Size}

To further optimize our network's performance and efficiency, we evaluate the effect of different convolutional kernel sizes within the attention modules. We experiment with kernel sizes of $1 \times 1$, $3 \times 3$, $5 \times 5$, and $7 \times 7$ on the Cityscapes validation set, with the results summarized in \autoref{ablation2}. The results demonstrate that increasing the kernel size generally leads to marginal improvements in segmentation accuracy, with both $5 \times 5$ and $7 \times 7$ kernels achieving the highest mean Intersection-over-Union (mIoU) of 79.5\%. However, the computational cost, measured in GFLOPs, increases notably with larger kernels—from 118.2 GFLOPs for $1 \times 1$ kernels up to 136.1 GFLOPs for $7 \times 7$ kernels. While the $7 \times 7$ kernel offers the best accuracy, the gain over the $5 \times 5$ kernel is minimal, suggesting diminishing returns as kernel size increases. On the other hand, the $1 \times 1$ kernel provides the lowest computational overhead but with slightly reduced accuracy. Balancing accuracy and efficiency, the $3 \times 3$ kernel size is a reasonable compromise, delivering near-peak performance (79.4\% mIoU) with moderate computational demands (120.8 GFLOPs). These observations suggest that moderate kernel sizes are preferable in practice, especially for real-time segmentation tasks where computational resources are constrained. 

\begin{table}[h]
  \caption{Ablation study on kernel size k in our network using Cityscapes validation set. Best results are highlighted in bold.}
  \label{ablation2}
  \centering
  \begin{tabular}{lll}
  \toprule
Method &mIoU (\%) &GFlops\\
   \midrule
k = 1 &79.2 &\textbf{118.2}\\
k = 3 &79.4 &120.8\\
k = 5 &79.5 &127.5\\
k = 7 &\textbf{79.5} &136.1\\
    \bottomrule
  \end{tabular}
\end{table}

\subsection{Ablation Study on Dilation Rates}

In our network, we employ an increasing sequence of dilation rates \(\{0, 1, 2, 3\}\) within the convolutional layers to capture multi-scale contextual information effectively. To investigate the impact of different dilation rates on the network’s performance, speed, and parameter count, we conducted a detailed ablation study on the Cityscapes validation set. The results are summarized in \autoref{ablation3}. As shown in the table, incorporating larger dilation rates progressively improves segmentation accuracy, with the highest mean Intersection-over-Union (mIoU) of 80.2\% achieved using a dilation rate of 3. This improvement is attributed to the enlarged receptive field, which allows the model to better capture global context and finer spatial details. However, the increase in dilation rate also leads to a rise in the number of model parameters—from 0.76 million for the baseline model without dilation up to 0.90 million for the model with dilation rate 3—and a corresponding decrease in inference speed (FPS). Specifically, the model with dilation rate 3 achieves an FPS of 88.6, which is lower than the 118.2 FPS of the baseline model. Interestingly, removing dilation altogether reduces the mIoU to 77.4\% but improves the FPS slightly to 121.3, indicating a trade-off between accuracy and speed. Overall, these results highlight the balance between capturing richer contextual information via dilation and maintaining real-time performance. The dilation rate of 2 offers a favorable compromise with an mIoU of 80.1\%, 102.7 FPS, and 0.84 million parameters.

\begin{table}
  \caption{Ablation study results on Cityscapes validation set using different dilation rates. FPS are estimated for an input image of resolution
of $512\times 1024$. Best results are highlighted in bold. }
  \label{ablation3}
  \centering
  \begin{tabular}{llll}
  \toprule
Model &mIoU (\%) &FPS &Parameters (M)\\
   \midrule
AASeg-baseline &79.2 &118.2 &\textbf{0.76}\\
AASeg-w/o dilation &77.4 &\textbf{121.3} &0.80\\
AASeg-(r=1) &79.8 &115.5 &0.79\\
AASeg-(r=2) &80.1 &102.7 &0.84\\
AASeg-(r=3) &\textbf{80.2} &88.6 &0.90\\
    \bottomrule
  \end{tabular}
\end{table}

\section{Conclusions}

In this paper, we introduced Attention Aware Network (AASeg), a novel architecture designed for real-time semantic segmentation. By integrating both Spatial Attention (SA) and Channel Attention (CA) modules, our network effectively enhances feature representations of objects without incurring additional computational overhead. The combined spatial-channel attention, together with the Multi Scale Context (MSC) module, enables robust and discriminative feature extraction for both targets and backgrounds. We extensively evaluated AASeg on multiple challenging benchmarks, including Cityscapes, ADE20K, and CamVid datasets. The results demonstrate that our approach achieves a strong balance between inference speed and segmentation accuracy, competitive with existing state-of-the-art methods. For future work, we plan to extend the multi-scale attention mechanisms within AASeg to tackle more complex tasks such as instance segmentation, aiming to further improve the precision and versatility of the model.

\nocite{*}
{
    \small
    \bibliographystyle{ieeenat_fullname}
    \bibliography{main}
}

\end{document}